\documentclass{article}
\usepackage{ijcai19}

\usepackage{times}
\usepackage{soul}
\usepackage{url}
\usepackage[hidelinks]{hyperref}
\usepackage[utf8]{inputenc}
\usepackage[small]{caption}
\usepackage{graphicx}
\usepackage{amsmath}
\usepackage{booktabs}
\usepackage{algorithm}
\usepackage{algorithmic}
\title{System Demo for Transfer Learning across Vision and Text \\using Domain Specific CNN Accelerator for On-Device NLP Applications}

\author{
Baohua Sun$^1$
\and
Lin Yang$^1$\and
Michael Lin$^1$\and
Wenhan Zhang$^1$\and
\\Patrick Dong$^1$\and
Charles Young$^1$\And
Jason Dong$^1$
\affiliations
$^1$Gyrfalcon Technology Inc.\\
\emails
\{baohua.sun\}@gyrfalcontech.com,
}

\begin{document}

\maketitle

\begin{abstract}
Power-efficient CNN Domain Specific Accelerator (CNN-DSA) chips are currently available for wide use in mobile devices. These chips are mainly used in computer vision applications. However, the recent work of Super Characters method for text classification and sentiment analysis tasks using two-dimensional CNN models has also achieved state-of-the-art results through the method of transfer learning from vision to text. In this paper, we implemented the text classification and sentiment analysis applications on mobile devices using CNN-DSA chips. Compact network representations using one-bit and three-bits precision for coefficients and five-bits for activations are used in the CNN-DSA chip with power consumption less than 300mW. For edge devices under memory and compute constraints, the network is further compressed by approximating the external Fully Connected (FC) layers within the CNN-DSA chip. At the workshop, we have two system demonstrations for NLP tasks. The first demo classifies the input English Wikipedia sentence into one of the 14 ontologies. The second demo classifies the Chinese online-shopping review into positive or negative.
\end{abstract}

\section{Introduction}
\begin{figure}[ht]
\vskip 0.2in
\begin{center}
\centerline{\includegraphics[width=0.8\columnwidth]{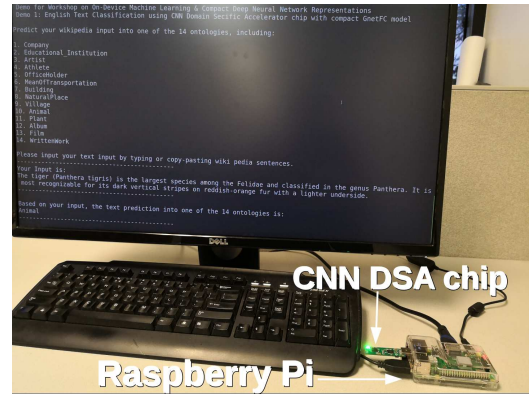}}
\caption{Efficient On-Device Natural Language Processing system demonstration. The CNN-DSA chip is connected to Raspberry Pi through the USB interface. Keyboard sends the typing text input to Raspberry Pi through USB. A monitor is connected to Raspberry Pi through HDMI for display. On the monitor, it shows the introduction for the demo (zoom in to see details). There are two demos. The first demo classifies the input English Wikipedia sentence into one of the 14 ontologies. The second demo classifies the Chinese online-shopping review into positive or negative.}
\label{SystemDemo}
\end{center}
\vskip -0.2in
\end{figure}

\begin{figure*}[ht]
\begin{center}
\centerline{\includegraphics[width=0.9\linewidth]{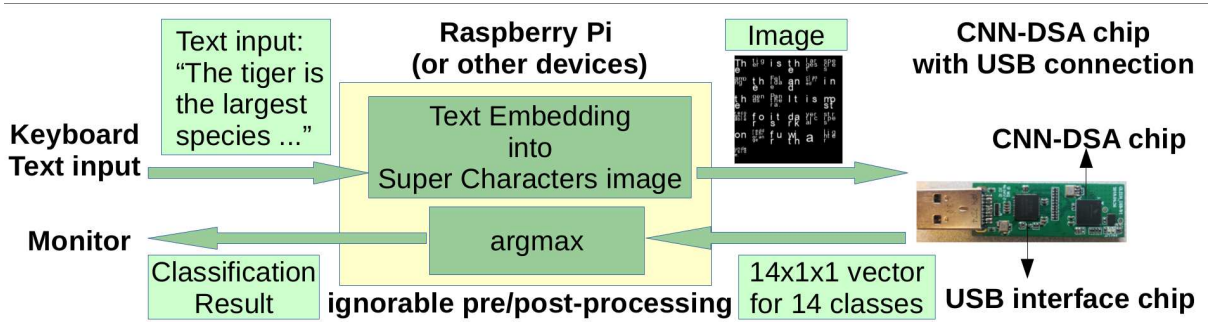}}
\caption{Data flow for the system demonstration of efficient on-device NLP using Super Characters method and CNN-DSA chip.}
\label{DataFlow}
\end{center}
\end{figure*}
Power-efficient CNN Domain Specific Accelerator (CNN-DSA) chips are currently available for wide use. Sun et al.~\shortcite{sun2018ultra,sun2018mram} designed a two-dimensional CNN-DSA accelerator which achieved a power consumption of less than 300mW and an ultra power-efficiency of 9.3TOPS/Watt. All the processing is in internal memory instead of external DRAM. Demos on mobile and embedded systems show its applications in real-world implementations. The 28nm CNN-DSA accelerator attains a 140fps for 224x224 RGB image inputs at an accuracy comparable to that of the VGG~\cite{simonyan2014very}.

For Natural Language Processing tasks, RNN and LSTM models~\cite{tang2015document,lai2015recurrent} are widely used, which are different network architectures from the two-dimensional CNN. However, the recent work of Super Characters method~\cite{sun2018super} using two-dimensional word embedding achieved state-of-the-art result in text classification and sentiment analysis tasks, showcasing the promise of this new approach. The Super Characters method is a two-step method. In the first step, the characters of the input text are drawn onto a blank image, so that an image of the text is generated with each of its characters embedded by the pixel values in the two-dimensional space. The resulting image is called the Super Characters image. In the second step, the generated Super Characters image is fed into a two-dimensional CNN model for classification. The two-dimensional CNN model is trained for the text classification task through the method of Transfer Learning, which finetunes the pretrained models on a large image dataset, e.g. ImageNet~\cite{imagenet_cvpr09}, with the labeled Super Characters images for the text classification task.

The follow-up works using the two-dimensional word embedding also show the effectiveness of this method in other applications. The SuperTML method~\cite{sun2019supertml} applies the two-dimensional word embedding to structured tabular data machine learning. Similar to the Super Characters method, it first prints the value of each attribute into non-overlapped segmentation of the image, and then feed the image into two-dimensional CNN model for classification. The experimental results also shows state-of-the-art reults on well-known data sets including Kaggle~\cite{Strata2017} and UCI Machine Learning Repository~\cite{UCIMachineLearningRep}. Othe applications of the two-dimensional word embedding includes dialogue generation for the chatbots~\cite{sun2019superchat} and image captioning~\cite{sun2019supercaptioning}. Experimental results show that high quality responses and captions are generated using the method of two-dimensional word embedding.

In this paper, we implemented NLP applications on mobile devices using the Super Characters method on a CNN-DSA chip as shown in Figure~\ref{SystemDemo}. It takes arbitrary text input from keyboard connecting to a mobile device (e.g. Raspberry Pi). And then the text is pre-processed into a Super Characters image and sent to the CNN-DSA chip to classify. After post-processing at the mobile device, the final result will be displayed on the monitor.

\section{System Design and Data Flow}
As shown in Figure~\ref{DataFlow}, the keyboard text input is pre-processed by the Raspberry Pi (or other mobile/embedded devices) to convert into a Super Characters image. This pre-processing is only a memory-write operation, which requires negligible computation and memory resources.

\begin{figure}[ht]
\vskip 0.2in
\begin{center}
\centerline{\includegraphics[width=0.5\columnwidth]{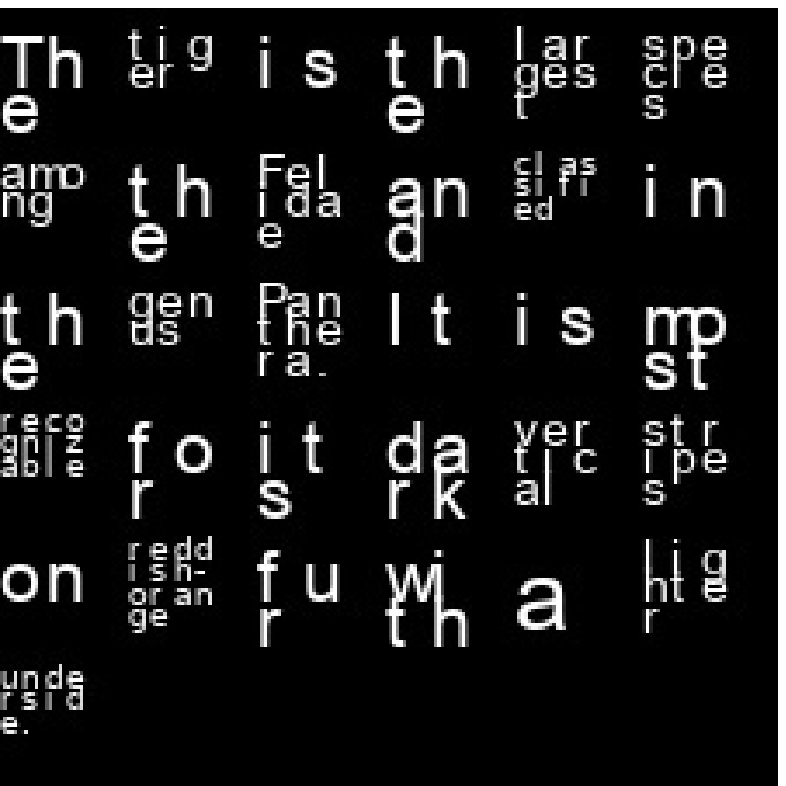}}
\caption{An example for Squared English Word (SEW) method. The two-dimensional embedding in this image corresponds to the text of "The tiger is the largest species among the Felidae and classified in the genus Panthera. It is most recognizable for its dark vertical stripes on reddish-orange fur with a lighter underside."}
\label{SEW}
\end{center}
\vskip -0.2in
\end{figure}

The Super Characters~\cite{sun2018super} method works well for Asian languages which has characters in squared shapes, such as Chinese, Japanese, and Korean. These glyphs are easier for CNN models to recognize than Latin languages such as English, which is alphabets-based in a rectangular shape and may have to break the words at line-changing. To improve the performance for English, a method of Squared English Word (SEW) is proposed~\cite{sun2019squared}. The intuition of the SEW method is to extend the original idea of Super Characters by preprocessing each English word into a squared glyph, just like Asian characters. To avoid information loss, the preprocessing should be a one-to-one mapping, i.e. each original English word can be recovered from the converted squared glyph. Figure~\ref{SEW} shows an example of this method.
Basically, each word takes the same size of a square space $l$x$l$. Words with longer alphabets will have smaller space for each alphabet. Within the $l$x$l$ space, the word with $N$ alphabets will have each of its alpha in the square area of $\{l/ceil[sqrt(N)]\}^2$, where $sqrt(.)$ stands for square root, and $ceil[.]$ is rounding to the top. 

The CNN-DSA chip receives the Super Characters image through the USB connection to the mobile device. It outputs the classification scores for the 14 classes in the Wikipedia text classification demo. The classification scores mean the probabilities for classification but before softmax. The mobile device only calculates the argmax to display final classification result on the monitor, which is also negligible computations. The CNN-DSA chip completes the complex CNN computations with low power less than 300mW. 
\section{Compact Network Representations for Efficient Inference}
\subsection{Approximating FC layers for On-Device Applications under Memory and Computation Constraints}
The CNN-DSA chip is a fast and low-power coprocessor. However, it does not directly support inner-product operations of the FC layers. It only supports 3x3 convolution, Relu, and max pooling. If the FC layers are executed on the mobile device, there will be increasing requirements for memory, computation, and storage for the FC coefficients. And it will also spend more interface time with the CNN-DSA chip for transmitting the activation map from the chip, and also cost relative high power consumption for the mobile device to execute the inner-product operations.

\begin{figure}[ht]
\vskip 0.2in
\begin{center}
\centerline{\includegraphics[width=\columnwidth]{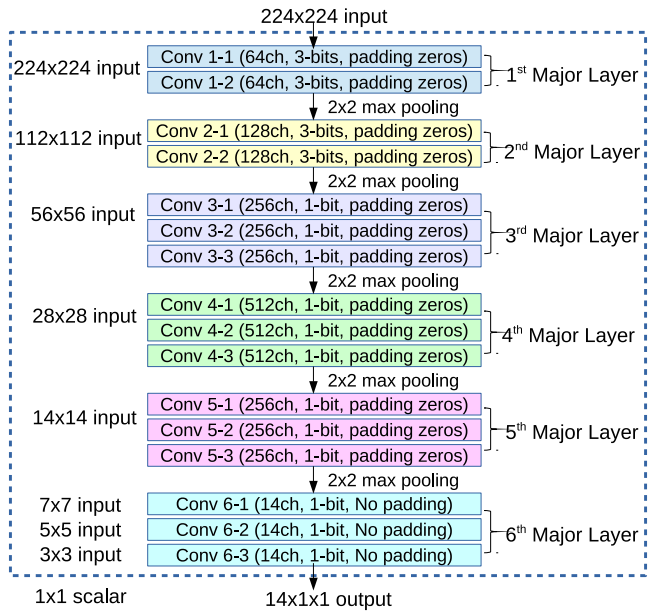}}
\caption{Model architecture. The input is of size 224x224 with multiple channels, and the output is of size 14x1x1. The architecture within blue dashed square is the model loaded into the CNN-DSA chip. }
\label{ModelArch}
\end{center}
\vskip -0.2in
\end{figure}

In order to address this problem, we proposed the GnetFC model (GTI-net with FC layers approximated by convolution layers), which approximates the FC layers using multiple layers of 3x3 convolutions. This is done by adding a sixth major layer with three sub-layers as shown in Figure~\ref{ModelArch}. 
The model is similar to VGG architecture except that it has six major layers, and the channels in the fifth major layer is reduced to 256 from the original 512 in order to save memory for the sixth layer due to the limitation of the on-chip memory. The sub-layers in each major layer has the same color. Each sub-layer name is followed by the detailed information in brackets, indicating the number of channels, bits-precision, and padding. The first five major layers has zero paddings at the image edge by one-pixel. But the sixth major layer has no padding for the three sublayers, which reduces the activation map from 7x7 through 5x5 and 3x3 and finally to 1x1. The output is of size 14x1x1, which is equal to an array of 14 scalars. The final classification result can be simply obtained by an argmax operation on the 14 scalars. This reduces the system memory footprint on the mobile device and accelerate the inference speed.

\subsection{Low-precision Inference in the Chip}
The memory of the CNN-DSA chip is built within the accelerator, so it is very power-efficient without wasting the energy for moving the bits from external DDR into internal SRAM. Thus the on-chip memory is very limited, which supports maximum 9MB for coefficients and activation map. As shown in Figure~\ref{ModelArch}, the first two major layers uses 3-bits precision and the other four major layers uses 1-bit precision. All activations are presented by 5-bits in order to save on-chip data memory. The representation mechanism inside the accelerator supports up to four times compression with the 1-bit precision, and two times compression with the 3-bits precision. Due to the high compression rate, the convolutional layers in VGG16 with 58.9MB coefficients in floating precision could be compressed into only about 5.5MB within the chip. This is a more than 10x compression of the convolution layers. This compact representation has been proved to be successful on ImageNet~\cite{imagenet_cvpr09} standard training and testing data and achieved the same level of accuracy as floating point models with 71\% Top1 accuracy. The compact CNN representation without accuracy loss is because of the redundancy in the original network. 

To efficiently use the on-chip memory, the model coefficients from the third major layers are only using 1-bit precision. For the first two major layers, 3-bits model coefficients are used as fine-grained filters from the original input image. And the cost on memory is only a quarter for the first major layer and a half for the second major layer if using the same 3-bits precision.

The total model size is 2.8MB, which is more than 200x compression from the original VGG model with FC layers. It completes all the convolution and FC processing within the CNN-DSA chip for the classification task with little accuracy drop. The GnetFC model on the CNN-DSA chip on the Wikipedia demo obtains an accuracy of 97.4\%, while the number for the original VGG model is 97.6\%. The accuracy drop is mainly brought by the approximation in GnetFC model, and also partially because of the bit-precision compression. The accuracy drop is very little, but the savings on power consumption and increasing on the inference speed is significant. It consumes less than 300mW on the CNN-DSA chip, and the power for pre/post-processing is negligible. The CNN-DSA chip processing time is 15ms, and the pre-processing time on mobile device is about 6ms. The time for post-processing is negligible, so the total text classification time is 21ms. It can process nearly 50 sentences in one second, which satisfies more than real-time requirement for NLP applications. 
\begin{table*}[t!]
\begin{center}
\begin{tabular}{|c|c|c|c|c|c|c|}
\hline Applications&Dataset& Language&Classes&Train&Test&On-chip Accuracy\\ \hline
Ontology Classification&DBpedia& English&14&560,000&70,000&97.4\%\\ \hline
Sentiment Classification&JD binary&Chinese&2&4,000,000&360,000&89.2\% \\
\hline
\end{tabular}
\end{center}
\caption{\label{tableExperiments} Data sets used in the experiments and experimental results.}
\end{table*}

\section{Experiments}

The data sets used and experimental results are shown in Table~\ref{tableExperiments}. For the application of Ontologies Classification for English inputs, the DBpedia data set~\cite{zhang2015character} is used. It classifies the English Wikipedia sentence input into 14 ontologies. Each ontology has 40,000 labeled text in training and 5,000 in testing. We use the SEW method and GnetFC model. The on-chip accuracy for the testing data set is 97.4\%. For the application of Sentiment Classification for Chinese inputs, the JD binary data set~\cite{zhang2017encoding} is used. It classifies the Chinese review for online-shopping into positive and negative. Each sentiment has 2,000,000 labeled text in training and 180,000 in testing. The original Super Characters method is used for the two-dimensional embedding because the input Chinese is already square-shaped glyph. The on-chip accuracy for the testing data set is 89.2\%.


\section{Conclusion}
We implemented efficient on-device NLP applications on a 300mW CNN-DSA chip by employing the two-dimensional embedding used in the Super Characters method. The two-dimensional embedding converts text into images, which is then fed into CNN-DSA chip for two-dimensional CNN computation. The demonstration system minimizes the power consumption of CNN for text classification, with less than 0.2\% accuracy drop from the original VGG model. The potential use cases for this demo system could be the intension recognition in a local-processing smart speaker or Chatbot.

\bibliographystyle{named}
\bibliography{ijcai19}

\end{document}